\documentclass[conference]{IEEEtran}
\IEEEoverridecommandlockouts

\usepackage{newtxtext}
\usepackage{CJKutf8}

\usepackage{amsmath,amssymb}
\usepackage{booktabs}
\usepackage{tabularx}
\usepackage{array}
\usepackage{multirow}
\usepackage{adjustbox}
\usepackage{graphicx}
\usepackage{xcolor}
\usepackage{tikz}
\usetikzlibrary{arrows.meta,decorations.pathreplacing}
\usepackage{url}
\usepackage[hidelinks]{hyperref}

\newcolumntype{Y}{>{\raggedright\arraybackslash}X}

\title{Active Perception for Embodied Disambiguation}
\author{Yiwei Liu and Luwei Yang\thanks{This work was supported by the Shenzhen Research Institute of Big Data (SRIBD) under Grant J00220250001. (Corresponding author: Luwei Yang.)\par Yiwei Liu is with the School of Science and Engineering, The Chinese University of Hong Kong, Shenzhen, Guangdong 518172, China (e-mail: 226015067@link.cuhk.edu.cn).\par Luwei Yang is with the Shenzhen Research Institute of Big Data, Shenzhen 518172, China (e-mail: yangluwei@sribd.cn).}}

\begin{document}
\begin{CJK*}{UTF8}{gbsn}
\maketitle

\begin{abstract}
Natural language provides robots with a flexible task interface, but target ambiguity in embodied environments arises not only from user intent; it can also result from missing task-relevant physical evidence in the current observation. Existing interactive disambiguation methods primarily obtain additional information by asking the user, whereas occlusion, restricted viewpoints, unreadable text, and unobserved targets require the robot to actively change its observation. We propose an active-perception framework for embodied target disambiguation that uses active observation as the backbone for information acquisition and uses a vision-language model to decide, on the basis of accumulated visual evidence and interaction information, whether to continue observing, request clarification, or complete target selection. Active observation can both directly recover missing discriminative evidence and reveal object names, labels, and semantic attributes, thereby improving user clarification when it remains necessary. Real-robot experiments show that the framework combines physical information acquisition and user-intent clarification within a unified embodied disambiguation process.
\end{abstract}

\section{Introduction}
\label{sec:introduction}

Recent advances in vision-language models have substantially strengthened the joint understanding of open-vocabulary visual content and natural-language instructions \cite{radford2021clip,alayrac2022flamingo,li2023blip2,liu2023llava}. At the same time, large-scale robot learning and vision--language--action models have further moved natural language from a perception interface into physical interaction, enabling robots to interpret scenes, select objects, and execute tasks from high-level linguistic descriptions \cite{ichter2023saycan,driess2023palme,brohan2022rt1,zitkovich2023rt2}. These advances have made natural language an increasingly important interface connecting user intent to robot behavior in embodied intelligent systems.

However, once natural language enters a physical environment, understanding an instruction does not imply that the target can be uniquely determined. Users may provide incomplete descriptions, multiple objects may simultaneously satisfy the current linguistic conditions, and the robot's visual input may be insufficient for reliable grounding. Interactive visual grounding and robotic target disambiguation therefore introduce user clarification: when current information is insufficient, the robot asks for additional information and progressively determines the target from the user's response \cite{shridhar2018interactive,shridhar2020ingress,mo2023seeask}. Subsequent research has further considered when to ask, how to formulate questions, and how to handle freer-form user responses \cite{ren2023knowno,park2024clara,kang2024prograsp}. These questions form the main trajectory of existing embodied disambiguation research and are discussed further in Section~II.

This interaction paradigm places the primary source of additional information on the user, but ambiguity in a real robot can also arise from incomplete observation. A key surface of an object may be occluded, target text may be unreadable because of distance or viewpoint, the camera may not see the side that determines object identity, or the requested target may not have entered the current useful observation range at all. A vision-language model can reason over available visual information, but it cannot generate physical evidence that is absent from a fixed image. As embodied multimodal models and robot foundation models increasingly emphasize closed-loop connections among vision, language, and physical action, whether a robot can use its own motion to actively change the information it receives becomes a natural next question \cite{openx2024rtx,kim2024openvla,octo2024generalist}.

This gives target disambiguation two different but related forms of missing information. The first is observation incompleteness: information required to solve the task exists in the physical world but has not yet been included in the current observation. The second is semantic or intent ambiguity: the scene has been sufficiently observed, but the user's preference needed to determine the target remains unknown. The two require different information sources. The former can be addressed by changing the viewpoint and observing the world again to obtain new physical evidence, whereas the latter requires the user to provide intent information that cannot be directly observed. Existing vision-language robotics research has demonstrated the potential of connecting multimodal reasoning with physical interaction, but how active observation and user clarification should jointly participate in information acquisition for interactive target disambiguation remains underexplored \cite{driess2023palme,zitkovich2023rt2,kim2024openvla}.

This paper studies this problem and proposes an embodied disambiguation framework with active observation as its information backbone. The vision-language model reasons over the user instruction, current and previously acquired visual observations, and available clarification information. When physical evidence is insufficient, the robot changes the viewpoint of its eye-in-hand camera and acquires a new RGB-D observation; after the new visual evidence returns to the model, it can trigger further observation or support target selection. If the scene has a sufficiently complete visual representation but the user's intent remains unresolved, the robot asks a clarification question grounded in the names, labels, and object attributes acquired through active observation. Active observation and user questioning are therefore not two independent remedial actions: a new observation first changes the robot's understanding of the physical scene and then changes the semantic basis available for subsequent clarification.

Real-robot experiments further examine this mechanism in three complementary situations: recovering discriminative evidence missing from the current local observation from new viewpoints, using actively acquired semantic attributes to improve clarification when user intent still requires confirmation, and changing the observed region to discover targets outside the initial useful view. The results show that active perception can not only directly resolve ambiguity caused by insufficient visual information, but can also provide a more complete physical grounding for language interaction when clarification remains necessary.

The main contributions of this paper are as follows:
\begin{itemize}
  \item We characterize embodied target disambiguation from the perspective of information sources, distinguishing observation incompleteness from semantic or intent ambiguity and thereby introducing active physical information acquisition into an interactive disambiguation problem centered on user clarification.
  \item We propose a vision-language disambiguation framework with active observation as its information backbone, allowing the robot to continuously acquire new physical evidence and grounding subsequent clarification and target selection in updated visual evidence.
  \item We validate through real-robot experiments the role of active perception in recovering missing discriminative evidence, improving the semantic basis of user clarification, and searching for targets outside the current useful observation.
\end{itemize}

\section{Related Work}
\label{sec:related}

\subsection{Interactive Disambiguation for Robotic Manipulation}
\label{sec:rw-interactive}

Interactive disambiguation for robots has developed along a clear question-centered trajectory.  Early work on interactive visual grounding showed that a referring expression need not be resolved in a single pass: when several image regions remain compatible with the language, the robot can ask the person for additional information \cite{shridhar2018interactive}.  INGRESS made this interaction more systematic by connecting candidate generation, relational grounding, and iterative, object-specific clarification in a robot manipulation system \cite{shridhar2020ingress}.  The important shift was from one-shot grounding to a dialogue in which the robot can expose the current ambiguity and obtain a correcting response.

The next step brought this interaction into cluttered manipulation.  INVIGORATE integrated visual grounding, blocking relationships, question generation, and grasping so that ambiguity resolution was part of a physical task rather than an isolated language benchmark \cite{zhang2021invigorate}.  Attribute-guided disambiguation then made the question itself a decision variable: instead of asking only for a generic confirmation, the robot can select an attribute question whose answers separate the remaining candidates \cite{yang2022attribute}.  This line of work therefore moved from whether the robot can ask to what it should ask.

SeeAsk extended the setting to open-world interactive disambiguation, with open-set objects and open-vocabulary interaction in real robotic grasping \cite{mo2023seeask}.  SeeAsk provides the two fixed-view asking strategies used as baselines in the present study: a greedy strategy confirms one currently preferred candidate at a time, whereas a static strategy asks from attributes available before any active camera motion.

The research question subsequently broadened from what to ask to when asking is warranted.  KnowNo addressed uncertainty alignment for language-model planners, giving the robot a calibrated basis for deciding when its current decision is insufficient for reliable execution \cite{ren2023knowno}.  CLARA made command-state recognition explicit by distinguishing clear, ambiguous, and infeasible commands before deciding whether to interact with the user \cite{park2024clara}.  These works establish that clarification is neither an automatic response to every instruction nor merely a final language-generation step.

The interaction channel itself has also become less rigid.  PROGrasp studied pragmatic human--robot communication in which the robot interprets intention-oriented and freer-form responses rather than assuming a fixed ``yes/no'' or option-ID protocol \cite{kang2024prograsp}.  More recently, AmbResVLM placed modern vision--language reasoning inside a robotic ambiguity-resolution pipeline, combining ambiguity detection, clarification generation, answer interpretation, and robotic validation \cite{chisari2025ambres}.  Taken together, these works progressively address whether ambiguity exists, what to ask, when to ask, and how to interpret the answer.

\subsection{From Asking for Information to Acquiring It from the World}
\label{sec:rw-world}

The common information channel in this line of work is the human response.  The robot observes a scene, forms a candidate hypothesis, and requests a linguistic distinction from the user \cite{shridhar2018interactive,shridhar2020ingress,zhang2021invigorate,mo2023seeask}.  This is appropriate when the missing variable is a preference or intention: vision can enumerate the available objects, but it cannot determine which medicine a person wants for an unstated reason.  Pragmatic interaction makes the same point from the opposite direction: a user may answer with an intention-bearing phrase rather than a constrained label, and the robot must interpret that answer \cite{kang2024prograsp}.

Not every ambiguity, however, is generated by language.  The distinguishing attribute may be occluded, text may be unreadable at the current viewpoint, a relevant surface may face away from the camera, or the requested target may lie outside the useful local observation.  In these cases, language reasoning cannot create the missing visual evidence.  Asking the user may reduce uncertainty among already observed alternatives, but it cannot reveal an alternative that has never entered the robot's observation.

This motivates a complementary information channel: embodied visual information acquisition.  The robot can first change what it sees, obtain a fresh RGB-D observation, and only then decide whether the remaining uncertainty belongs to the world or to the user's intent.  Modern VLM-based ambiguity resolution provides the language and visual reasoning substrate for this interaction, while embodied observation adds information unavailable from a fixed image \cite{ren2023knowno,park2024clara,chisari2025ambres}.  The central question is therefore not only what the robot should ask or when it should ask, but whether the missing information should first be acquired from the user or from the physical world.

\section{Problem Formulation}
\label{sec:formulation}

\subsection{Embodied Target Disambiguation}
\label{sec:formulation-basic}

Let $u$ denote the user's instruction, $o$ a visual observation, and $g^\star$ the user's intended physical target.  The robot resolves $g^\star$ by combining visual evidence with information obtained through embodied interaction.  The VLM makes a high-level decision $\alpha \in \{\mathsf{O},\mathsf{Q},\mathsf{S}\}$, where $\mathsf{O}$ acquires another physical camera observation, $\mathsf{Q}$ queries the user for clarification, and $\mathsf{S}$ commits to the resolved physical target.  These decisions are not independent information channels: observation establishes the physical evidence on which clarification and target commitment operate.

\subsection{Two Sources of Embodied Ambiguity}
\label{sec:taxonomy}

Two sources of embodied ambiguity are distinguished according to where the missing task-relevant information resides.

\paragraph{Observation incompleteness}
The current observation does not provide all task-relevant physical evidence required to resolve the target.  Let $\mathcal{C}(o)$ denote the candidate hypothesis supported by observation $o$.  An incomplete observation may either provide insufficient evidence to distinguish candidates already in $\mathcal{C}(o)$---for example, when a target surface is occluded, the viewpoint is insufficient, printed text is too small or unreadable, or a relevant side is not exposed---or omit a relevant physical target from $\mathcal{C}(o)$ altogether.  A new observation can therefore improve evidence about an already observed target or expand the currently observed candidate set by inspecting another part of the workspace.

\paragraph{Semantic or intent ambiguity}
The environment may be sufficiently visible, while the instruction still fails to specify the user's preference.  An instruction can identify a class of objects without stating which functional or semantic property the user wants.  Active observation can improve the evidence available for asking, but it cannot replace the missing preference.

\section{Active Perception for Embodied Disambiguation}
\label{sec:method}

The process is organized around the acquisition and accumulation of embodied visual evidence.  The VLM reasons from the user instruction, the current camera view, previously acquired observations, and any clarification dialogue.  Active observation changes the physical evidence available to subsequent reasoning; after each fresh view, the VLM reassesses the scene and may continue observing, formulate a clarification grounded in the acquired evidence, or commit to a resolved target.

\begin{figure*}[t]
\centering
\resizebox{\textwidth}{!}{%

\definecolor{archReasonFill}{RGB}{237,244,253}
\definecolor{archReasonLine}{RGB}{35,72,132}
\definecolor{archObsFill}{RGB}{239,247,234}
\definecolor{archObsLine}{RGB}{75,119,58}
\definecolor{archClarFill}{RGB}{255,247,213}
\definecolor{archClarLine}{RGB}{181,137,18}
\definecolor{archSelectFill}{RGB}{255,241,231}
\definecolor{archSelectLine}{RGB}{232,113,53}
\definecolor{archPurple}{RGB}{105,42,163}

\tikzset{
  archbox/.style={rounded corners=4pt, line width=0.75pt, align=center, inner sep=1.5pt},
  archarrow/.style={-{Latex[length=2.6mm,width=1.8mm]}, line width=0.75pt},
  archlabel/.style={font=\small\itshape, align=center, text=black},
}

\begin{tikzpicture}[x=1cm,y=1cm,text=black]

\draw[archbox, fill=white, draw=black] (0.10,5.70) rectangle (3.10,6.80);
\node[font=\small\bfseries, align=center, text width=2.70cm] at (1.60,6.25) {User instruction $u$};
\draw[archbox, fill=white, draw=black] (0.10,3.60) rectangle (3.10,4.70);
\node[font=\small\bfseries, align=center, text width=2.70cm] at (1.60,4.15) {Initial RGB-D\\observation $o$};

\draw[archbox, fill=archReasonFill, draw=archReasonLine, line width=1pt]
  (3.95,4.05) rectangle (7.95,6.45);
\node[align=center, text width=3.55cm] at (5.95,5.25) {
  \textbf{VLM reasoning}\\[2pt]
  {\itshape uses current view,\\previous observations,\\and clarification dialogue}};

\draw[archbox, fill=archObsFill, draw=archObsLine, line width=1pt]
  (8.50,4.05) rectangle (11.00,6.45);
\node[align=center, text width=2.20cm] at (9.75,5.25) {
  {\bfseries\itshape $\alpha=\mathsf{O}$}\\[4pt]
  Active\\observation};

\draw[archbox, fill=archObsFill, draw=archObsLine, line width=1pt]
  (11.55,4.25) rectangle (14.15,6.25);
\node[align=center, text width=2.30cm] at (12.85,5.25) {Eye-in-hand\\camera changes\\viewpoint};

\draw[archbox, fill=archObsFill, draw=archObsLine, line width=1pt]
  (14.75,3.95) rectangle (19.25,6.55);
\node[font=\small\bfseries, align=center, text width=4.05cm] at (17.00,6.18) {Fresh RGB-D evidence};
\draw[archbox, dashed, fill=white, draw=archObsLine, line width=0.7pt]
  (15.15,5.10) rectangle (18.85,5.85);
\node[font=\footnotesize, align=center, text width=3.35cm] at (17.00,5.475) {reveal discriminative\\evidence};
\draw[archbox, dashed, fill=white, draw=archObsLine, line width=0.7pt]
  (15.15,4.20) rectangle (18.85,4.95);
\node[font=\footnotesize, align=center, text width=3.35cm] at (17.00,4.575) {inspect another region /\\unseen target};

\draw[archbox, fill=archReasonFill, draw=archReasonLine, line width=1pt]
  (19.95,4.15) rectangle (24.75,6.35);
\node[align=center, text width=4.25cm] at (22.35,5.25) {
  \textbf{Updated VLM reasoning}\\[3pt]
  {\itshape reassesses the scene with\\accumulated visual evidence}};

\draw[archbox, fill=archClarFill, draw=archClarLine, line width=1pt]
  (20.50,2.10) rectangle (23.95,3.65);
\node[font=\footnotesize, align=center, text width=3.10cm] at (22.225,2.875) {
  {\bfseries\itshape $\alpha=\mathsf{Q}$}\\[2pt]
  \mbox{Observation-grounded}\\[-1pt]clarification};
\draw[archbox, fill=white, draw=archPurple, line width=1pt]
  (20.15,0.35) rectangle (23.35,1.55);
\node[font=\small\bfseries, align=center, text width=2.85cm] at (21.75,0.95) {Human response};

\draw[archbox, fill=archSelectFill, draw=archSelectLine, line width=1pt]
  (25.20,4.15) rectangle (28.45,6.35);
\node[align=center, text width=2.90cm] at (26.825,5.25) {
  {\bfseries\itshape $\alpha=\mathsf{S}$}\\[4pt]
  Observation-grounded\\target selection};
\draw[archbox, fill=white, draw=black]
  (29.55,4.40) rectangle (32.60,6.10);
\node[font=\small\bfseries, align=center, text width=2.70cm] at (31.075,5.25) {Resolved\\physical target};

\draw (3.10,6.25) -- (3.45,6.25) -- (3.45,5.25);
\draw (3.10,4.15) -- (3.45,4.15) -- (3.45,5.25);
\draw[archarrow] (3.45,5.25) -- (3.95,5.25);
\draw[archarrow] (7.95,5.25) -- (8.50,5.25);
\draw[archarrow] (11.00,5.25) -- (11.55,5.25);
\draw[archarrow] (14.15,5.25) -- (14.75,5.25);
\draw[archarrow] (19.25,5.25) -- (19.95,5.25);
\draw[archarrow] (24.75,5.25) -- (25.20,5.25);
\draw[archarrow] (28.45,5.25) -- (29.55,5.25);

\draw[archarrow] (22.35,6.35) -- (22.35,8.35) -- (9.75,8.35) -- (9.75,6.45);
\node[font=\normalsize\bfseries\itshape] at (17.10,8.78) {$\alpha=\mathsf{O}$};

\draw[decorate, decoration={brace, amplitude=14pt, mirror}, draw=archReasonLine, line width=0.7pt]
  (8.65,3.88) -- (19.10,3.88);
\node[archlabel] at (14.35,2.90) {physical evidence acquisition};

\draw[archarrow] (22.35,4.15) -- (22.35,3.65);
\draw[archarrow] (22.225,2.10) -- (22.225,1.55);

\draw[-{Latex[length=2.6mm,width=1.8mm]}, draw=archPurple, line width=0.9pt]
  (23.35,0.95) -- (23.90,0.95)
  .. controls (24.35,0.95) and (24.35,1.35) .. (24.35,1.85)
  -- (24.35,4.05);
\node[archlabel, text width=2.55cm] at (25.45,1.75) {intent /\\preference\\information};
\node[archlabel, text width=2.85cm, align=left] at (26.35,3.10)
  {question grounded in\\names, labels, or\\visual attributes};

\end{tikzpicture}%
}
\caption{Overview of the active-perception-based embodied disambiguation framework.  Active observation provides the visual evidence used by subsequent clarification and target selection.}
\label{fig:architecture}
\end{figure*}
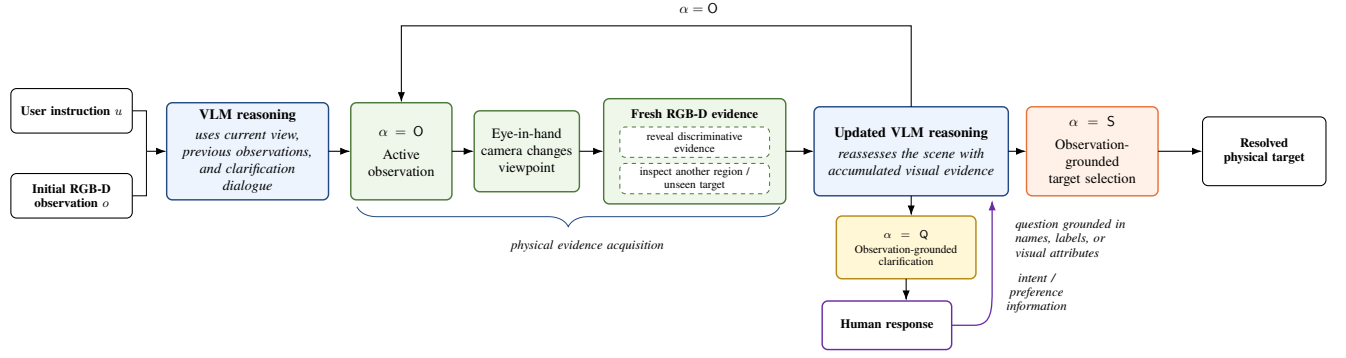

\subsection{Active Observation as the Information Backbone}
\label{sec:method-observation}

Active observation is the mechanism through which the robot changes the physical evidence available to its reasoning.  When the current view does not provide enough task-relevant information, the VLM can choose $\alpha=\mathsf{O}$; the eye-in-hand camera then changes viewpoint, acquires a fresh RGB-D observation, and returns it to the VLM.  After each fresh view, the VLM reassesses the scene and can choose another observation when the evidence remains incomplete, so $\mathsf{O}$ may be repeated.  New views can reveal an occluded surface, make small text readable, expose a label or another side of an object, identify a semantic attribute, inspect another tabletop region, or reveal a previously unseen target.  The information obtained through $\mathsf{O}$ forms the visual basis for the subsequent clarification or target-selection decision.

\subsection{Observation-Grounded Clarification}
\label{sec:method-clarification}

Clarification is generated from the visual understanding built through observation.  The initial view may support only a coarse description such as an object on the right with a colorful pattern, whereas an additional observation may expose a medicine name, printed package label, object identity, or another meaningful semantic attribute.  Active observation therefore does not merely precede clarification; it changes the semantic evidence from which the clarification is formed.  When the physical alternatives are sufficiently characterized but the user's preference remains unresolved, the VLM can choose $\alpha=\mathsf{Q}$ and use these visually grounded attributes to formulate a clarification question.  The user's response adds intent or preference information to the subsequent reasoning; if that response makes further visual evidence necessary, the updated reasoning can lead to another observation before target commitment.

\subsection{Observation-Grounded Target Selection}
\label{sec:method-selection}

Target selection is the terminal decision made from the evidence accumulated through observation and, when needed, clarification.  Once the accumulated visual evidence and any required clarification identify a target consistently, the VLM chooses $\alpha=\mathsf{S}$ and commits to that physical target.  Depending on the evidence, the information flow may contain repeated observations before selection, or observation followed by grounded clarification and then selection; after a user response, subsequent reasoning may also lead to another observation, another clarification, or target commitment.

\section{Real-Robot Experiments}
\label{sec:experiments}

\subsection{Platform and Protocol}
\label{sec:platform}

\begin{figure}[t]
\centering
\includegraphics[width=\columnwidth]{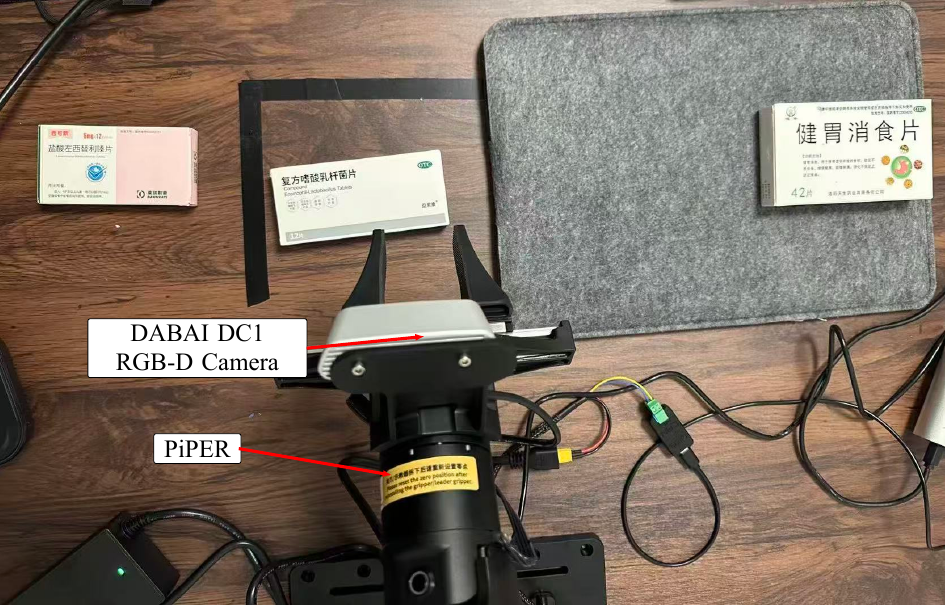}
\caption{Experimental setup of the PiPER platform with the eye-in-hand DABAI DC1 RGB-D camera.}
\label{fig:setup-photo}
\end{figure}

All matched real-robot experiments use a PiPER six-degree-of-freedom manipulator with a DABAI DC1 RGB-D camera in an eye-in-hand configuration.  The robot operates on real tabletop scenes and obtains synchronized RGB-D observations before and after active camera motion. The experimental setup is shown in Fig.~\ref{fig:setup-photo}.

Each matched comparison uses the same physical scene configuration, the same initial observation condition, and the same user instruction across methods.  All matched trials used Qwen3.7-Plus (model ID: qwen3.7-plus), accessed through an Alibaba Cloud Model Studio (Bailian) workspace in the China (Beijing, cn-beijing) region.  The protocol evaluates ambiguity resolution before target commitment.

\subsection{Baselines}
\label{sec:methods}

\textbf{Passive Direct Selection} uses the initial observation only, asks no question, and directly selects one candidate.

\textbf{Greedy Asking} is an adapted SeeAsk-style baseline \cite{mo2023seeask}.  Before target selection it does not move the camera.  It asks a confirmation-style question about one currently preferred candidate at a time and advances using the user's answer.

\textbf{Static Asking} is also an adapted SeeAsk-style baseline \cite{mo2023seeask}.  It does not move the camera before target selection and constructs a candidate-partition question using only attributes available in the frozen initial view.

\subsection{Matched Scenarios and Research Questions}
\label{sec:dataset}

The main comparison contains nine matched scenarios and four methods per scenario, for 36 real-robot trials.  Scene configuration, ambiguity regime, requested evidence, and outcomes are summarized in Table~\ref{tab:matched}.

The experiments address three operational research questions:
\begin{itemize}
  \item \textbf{RQ1 (A1--A5):} Can active perception recover discriminative visual evidence for a target already represented in the initial local observation?
  \item \textbf{RQ2 (A6--A7):} Can active perception provide better semantic grounding for a clarification that remains necessary?
  \item \textbf{RQ3 (A8--A9):} Can active perception expand the observed workspace when the requested target is absent from the initial candidate set?
\end{itemize}

\subsection{Evaluation}
\label{sec:evaluation}

Evaluation uses four complementary dimensions.  Target or ambiguity resolution records whether the user's target was resolved.  Grounding source distinguishes fresh visual evidence, user clarification, initial-view evidence, and unsupported selection.  Clarification interaction records the question count together with the recorded exchange, and active perception records the number of new viewpoints obtained before selection.  For semantic-intent scenarios A6 and A7, the intended target is determined by the recorded clarification response in each interactive trial.  Passive elicits no preference and is therefore reported as unsupported.

\subsection{Recovering Discriminative Evidence for Initially Observed Targets}
\label{sec:results-visual}

All five scenarios contain the requested target within the initial local scene, but the initial observation does not provide sufficient discriminative evidence for reliable resolution.  In A1--A3, the Proposed method obtained the occluded, fine-text, and volume-specific attributes from fresh views.  Passive was correct only in the first case and incorrect in the other two; Greedy and Static resolved the targets through user clarification without acquiring new visual observations.  In A4, the Proposed, Greedy, and Static methods resolved the requested book while Passive selected the other book.  In A5, all four methods resolved the requested box.  Table~\ref{tab:matched} records these outcomes and their evidence sources, which distinguish correctness from grounding source.

\subsection{Semantic Grounding for User Clarification}
\label{sec:results-clarification}

The semantic scenarios keep the physical scene visible while leaving the user's desired medicine underspecified.  Active observation changes the semantic attributes available to the question.  The recorded questions and user responses are consolidated in Table~\ref{tab:dialogues}, with original Chinese records separated from faithful English translations.

\paragraph{A6: active survey before asking.}
The active survey exposes package names and semantic attributes before clarification.  Proposed asks using medicine names, Greedy uses a binary appearance-based confirmation, and Static relies on fixed-view descriptions.  The user response resolves each interactive trial, whereas Passive asks no question and remains an unsupported direct selection.

\paragraph{A7: free-form answers matter.}
The free-form user answer explicitly names the digestive-aid medicine.  Proposed, Greedy, and Static all resolve that named target, while Passive remains an unsupported incorrect guess.  The A7 Greedy interaction exposes a limitation of candidate-specific binary confirmation: after the user explicitly named the desired medicine, the baseline repeated the same appearance-based confirmation instead of exploiting the richer semantic information already provided.  The target was eventually resolved after the user answered the repeated binary question with ``Yes,'' but the extra turn was unnecessary once the medicine name had already been supplied.  The complete multi-turn interaction is retained in Table~\ref{tab:dialogues}.

\subsection{Searching Beyond the Initial Observation}
\label{sec:results-search}

\paragraph{A8: search beyond the local observation.}
The Proposed method recognized that the requested medicine was absent from the current local observation, continued active search after the user reiterated the target, and selected it after a later fresh view.  The fixed-view methods remained confined to the observed alternatives.  The representative questions and responses are reported in Table~\ref{tab:dialogues}.

\paragraph{A9: farther out-of-view search.}
The Proposed method obtained a fresh view, read the requested label, and selected the target.  Passive, Greedy, and Static remained at zero pre-selection active moves and did not resolve the distant target.  Together, A8 and A9 show that changing the observation can expand the candidate hypothesis beyond the initially visible local set.

\begin{table*}[!t]
\centering
\caption{Matched real-robot scenarios and outcomes.  ``Obs.'' denotes the number of pre-selection active observations, and ``Q'' denotes the number of clarification questions.}
\label{tab:matched}
\scriptsize
\renewcommand{\arraystretch}{0.9}
\begin{tabularx}{\textwidth}{@{}l l >{\raggedright\arraybackslash}p{1.55cm} >{\raggedright\arraybackslash}p{2.05cm} l >{\raggedright\arraybackslash}p{1.75cm} >{\raggedright\arraybackslash}p{1.85cm} c c@{}}
\toprule
Scene & Regime & Scene setup & Task / requested evidence & Method & Resolution outcome & Grounding source & Obs. & Q\\
\midrule
\multirow{4}{*}{A1} & \multirow{4}{*}{\shortstack[l]{Observation\\incompleteness}} & \multirow{4}{1.55cm}{Top surface partly occluded} & \multirow{4}{2.05cm}{Letter ``A'' on box top} & Proposed & Correct & Fresh RGB-D & 3 & 0\\
& & & & Passive & Correct & Unsupported initial-view selection & 0 & 0\\
& & & & Greedy & Correct & User answer; fixed view & 0 & 1\\
& & & & Static & Correct & User answer; fixed view & 0 & 1\\
\multirow{4}{*}{A2} & \multirow{4}{*}{\shortstack[l]{Observation\\incompleteness}} & \multirow{4}{1.55cm}{Small text at initial distance} & \multirow{4}{2.05cm}{Book text ``美国'' (``USA'')} & Proposed & Correct & Fresh RGB-D & 3 & 0\\
& & & & Passive & Incorrect & Initial-view evidence & 0 & 0\\
& & & & Greedy & Correct & User answer; fixed view & 0 & 1\\
& & & & Static & Correct & User answer; fixed view & 0 & 1\\
\multirow{4}{*}{A3} & \multirow{4}{*}{\shortstack[l]{Observation\\incompleteness}} & \multirow{4}{1.55cm}{Two same-title books; volume marker} & \multirow{4}{2.05cm}{First volume of same-title books} & Proposed & Correct & Fresh RGB-D & 16 & 0\\
& & & & Passive & Incorrect & Unsupported initial-view selection & 0 & 0\\
& & & & Greedy & Correct & User answer; fixed view & 0 & 1\\
& & & & Static & Correct & User answer; fixed view & 0 & 1\\
\multirow{4}{*}{A4} & \multirow{4}{*}{\shortstack[l]{Observation\\incompleteness}} & \multirow{4}{1.55cm}{Two books; one trilogy title} & \multirow{4}{2.05cm}{Book ``地球往事三部曲'' (``Earth's Past trilogy'')} & Proposed & Correct & Fresh RGB-D & 1 & 0\\
& & & & Passive & Incorrect & Initial-view evidence & 0 & 0\\
& & & & Greedy & Correct & User answer; fixed view & 0 & 1\\
& & & & Static & Correct & User answer; fixed view & 0 & 1\\
\multirow{4}{*}{A5} & \multirow{4}{*}{\shortstack[l]{Observation\\incompleteness}} & \multirow{4}{1.55cm}{Two boxes; PLUS PEN S label} & \multirow{4}{2.05cm}{Box text ``PLUS PEN S''} & Proposed & Correct & Fresh RGB-D & 2 & 0\\
& & & & Passive & Correct & Initial-view evidence & 0 & 0\\
& & & & Greedy & Correct & User answer; fixed view & 0 & 1\\
& & & & Static & Correct & User answer; fixed view & 0 & 1\\
\multirow{4}{*}{A6} & \multirow{4}{*}{Semantic / intent} & \multirow{4}{1.55cm}{Three boxes; preference unspecified} & \multirow{4}{2.05cm}{Select among three medicine boxes} & Proposed & Correct (run-level) & Fresh RGB-D + user answer & 1 & 1\\
& & & & Passive & Unsupported selection & Initial-view evidence; no preference & 0 & 0\\
& & & & Greedy & Correct (run-level) & User answer; fixed view & 0 & 1\\
& & & & Static & Correct (run-level) & User answer; fixed view & 0 & 1\\
\multirow{4}{*}{A7} & \multirow{4}{*}{Semantic / intent} & \multirow{4}{1.55cm}{Three boxes; named preference} & \multirow{4}{2.05cm}{Select among three medicine boxes} & Proposed & Correct (run-level) & Fresh RGB-D + user answer & 2 & 1\\
& & & & Passive & Incorrect & Initial-view unsupported selection & 0 & 0\\
& & & & Greedy & Correct (run-level) & Free-form answers; fixed view & 0 & 2\\
& & & & Static & Correct (run-level) & Free-form answer; fixed view & 0 & 1\\
\multirow{4}{*}{A8} & \multirow{4}{*}{\shortstack[l]{Observation\\incompleteness}} & \multirow{4}{1.55cm}{Requested box outside local view} & \multirow{4}{2.05cm}{Box ``健胃消食片'' (``digestive-aid tablets'')} & Proposed & Correct & Fresh view after search & 3 & 1\\
& & & & Passive & Incorrect & Local initial candidate & 0 & 0\\
& & & & Greedy & Incorrect & Local candidate + answer & 0 & 1\\
& & & & Static & Incorrect & Local candidate + answer & 0 & 1\\
\multirow{4}{*}{A9} & \multirow{4}{*}{\shortstack[l]{Observation\\incompleteness}} & \multirow{4}{1.55cm}{Requested box farther out of view} & \multirow{4}{2.05cm}{Box ``复方嗜酸乳杆菌片'' (``probiotic tablets'')} & Proposed & Correct & Fresh view after search & 1 & 0\\
& & & & Passive & Incorrect & Local initial candidate & 0 & 0\\
& & & & Greedy & Incorrect & Local initial candidate & 0 & 0\\
& & & & Static & Incorrect & Local initial candidate & 0 & 0\\
\bottomrule
\end{tabularx}
\end{table*}

\begin{table*}[!t]
\centering
\caption{Representative Human--Robot Clarification Transcripts.  The Original Chinese Interaction column preserves the recorded Chinese exchange; the English Translation column gives a faithful English rendering.  Passive trials are omitted because they contain no clarification dialogue.}
\label{tab:dialogues}
\footnotesize
\setlength{\tabcolsep}{3pt}
\renewcommand{\arraystretch}{1.12}
\begin{tabularx}{\textwidth}{@{}p{0.07\textwidth}p{0.10\textwidth}Y Y@{}}
\toprule
Scene & Method & Original Chinese Interaction & English Translation\\
\midrule
A6 & Proposed & \textbf{Robot:} 桌上有三盒药：左边是粉色的‘盐酸左西替利嗪片’，中间是白色的‘复方嗜酸乳杆菌片’，右边是白色的‘健胃消食片’。请问您需要拿哪一盒？\par \textbf{User:} 我想要一个可以治疗肠胃的 & \textbf{Robot:} There are three medicine boxes on the table: the pink one on the left is Levocetirizine Hydrochloride Tablets, the white one in the middle is Compound Eosinophil-Lactobacillus Tablets, and the white one on the right is Jianwei Xiaoshi Digestive Tablets. Which one do you need?\par \textbf{User:} I want one that can treat stomach and intestinal problems.\\
A6 & Greedy & \textbf{Robot:} 你指的是右边带彩色图案的物体吗？\par \textbf{User:} 是 & \textbf{Robot:} Do you mean the object on the right with a colorful pattern?\par \textbf{User:} Yes.\\
A6 & Static & \textbf{Robot:} 你要左边白色的物体，还是右边带条纹侧面的物体或右边带彩色图案的物体？\par \textbf{User:} 右边带条纹侧面的物体 & \textbf{Robot:} Do you mean the white object on the left, the object on the right with a striped side, or the object on the right with a colorful pattern?\par \textbf{User:} The object on the right with a striped side.\\
A7 & Proposed & \textbf{Robot:} 请问您需要拿哪一盒药？是左边粉色包装的，中间白色带绿色条纹的，还是右边印有‘健胃消食片’字样的？\par \textbf{User:} 我想要健胃消食片 & \textbf{Robot:} Which box of medicine do you need: the one with pink packaging on the left, the one with white packaging and green stripes in the middle, or the one on the right printed with the words Jianwei Xiaoshi Digestive Tablets?\par \textbf{User:} I want Jianwei Xiaoshi Digestive Tablets.\\
A7 & Greedy & \textbf{Robot 1:} 你指的是右边带彩色图案的物体吗？\par \textbf{User 1:} 我要健胃消食片\par \textbf{Robot 2:} 你指的是右边带彩色图案的物体吗？\par \textbf{User 2:} 是 & \textbf{Robot 1:} Do you mean the object on the right with a colorful pattern?\par \textbf{User 1:} I want Jianwei Xiaoshi Digestive Tablets.\par \textbf{Robot 2:} Do you mean the object on the right with a colorful pattern?\par \textbf{User 2:} Yes.\\
A7 & Static & \textbf{Robot:} 你要左边白色的物体，还是右边黑色的物体或右边带彩色图案的物体？\par \textbf{User:} 我要健胃消食片 & \textbf{Robot:} Do you want the white object on the left, the black object on the right, or the object on the right with a colorful pattern?\par \textbf{User:} I want Jianwei Xiaoshi Digestive Tablets.\\
A8 & Proposed & \textbf{Robot:} 请问您需要的是左边粉色包装的‘盐酸左西替利嗪片’，还是右边白色包装的‘复方嗜酸乳杆菌片’？当前视野中未发现名为‘健胃消食片’的药盒。\par \textbf{User:} 我要健胃消食片 & \textbf{Robot:} Do you need the left pink package of Levocetirizine Hydrochloride Tablets or the right white package of Compound Eosinophil-Lactobacillus Tablets? The current view does not contain a medicine box named Jianwei Xiaoshi Digestive Tablets.\par \textbf{User:} I want Jianwei Xiaoshi Digestive Tablets.\\
A8 & Greedy & \textbf{Robot:} 你指的是白色药盒黑色的物体吗？\par \textbf{User:} 是 & \textbf{Robot:} Do you mean the black object that is a white medicine box?\par \textbf{User:} Yes.\\
A8 & Static & \textbf{Robot:} 你要左边白色的物体，还是Compound Paracetamol and Amantadine Hydrochloride Capsules Box黑色的物体？\par \textbf{User:} 左边白色的物体 & \textbf{Robot:} Do you want the white object on the left, or the black object identified as the Compound Paracetamol and Amantadine Hydrochloride Capsules box?\par \textbf{User:} The white object on the left.\\
\bottomrule
\end{tabularx}
\end{table*}

\section{Discussion}
\label{sec:discussion}

\subsection{Not Every Ambiguity Should Trigger a Question}
\label{sec:discussion-first-look}

The experiments show two complementary responses to ambiguity.  When the missing information is in the world---an occluded surface, unreadable text, or an unobserved region---active observation supplies it.  When the scene is sufficiently visible but the user preference remains unspecified, clarification supplies the missing intent.  The source of the uncertainty therefore determines the useful next action.

\subsection{Seeing and Asking Provide Different Information}
\label{sec:discussion-channels}

Active observation and clarification provide different information.  The world answers ``What is physically there?'' while the user answers ``What do I actually want?''  The two channels are complementary: visual evidence identifies available targets and user responses specify the intended one.

\subsection{Active Perception Can Improve the Language Interface}
\label{sec:discussion-interface}

The A6 and A7 traces show that active perception can improve a language interface even when it does not eliminate the need for a question.  The fresh view exposes package names, colors, stripes, graphics, and other attributes.  Those attributes support a more grounded question than a fixed-view positional proxy, and the user then contributes a preference or intention, including a free-form answer.  The A7 interaction further shows that a semantically informative user response can be wasted by a rigid appearance-based confirmation strategy, whereas visually grounded object identities allow the clarification process to operate at the semantic level of the task.  Physical information acquisition thus feeds into linguistic interaction; its value is not limited to object recognition.

\subsection{Scope of Pretrained Semantic Knowledge}

The experiments deliberately focus on tasks that cannot be resolved from the VLM's pretrained knowledge alone. The decisive information in the evaluated scenarios is specific to the current interaction, such as a letter physically attached to a particular box, a volume marker visible on one of two otherwise similar books, or whether a requested medicine box is present within the currently observed workspace. A VLM may already understand the meanings of these words and object categories, but such prior knowledge cannot determine which physical object in the current environment satisfies the instruction and therefore cannot replace direct observation of the scene. This experimental design intentionally separates online information acquisition from the model's pretrained semantic knowledge, allowing the evaluation to isolate when new physical evidence must be acquired from the world and when intent or preference information must instead be obtained from the user. Accordingly, the conclusions of this work primarily apply to embodied disambiguation problems in which the decisive task-specific information must be acquired during the current interaction. A different class of problems arises when active observation reveals only an intermediate semantic cue and the VLM subsequently uses pretrained knowledge to infer a relation between concepts, attributes, or labels before resolving the target. In such cases, the final decision depends jointly on acquired evidence and prior semantic knowledge, and the respective contributions can no longer be characterized solely in terms of information obtained from the world or the user. These cases are intentionally excluded from the present evaluation to preserve the interpretability of the studied information-acquisition problem. A natural extension is therefore to treat pretrained world knowledge explicitly as a third information resource and study how physical observation, user feedback, and prior knowledge should be combined during sequential embodied interaction.

\subsection{Consistency Between Physical Evidence and User Feedback}

This work primarily considers cases in which physical observations and user feedback are mutually consistent, allowing the two online information sources to assume relatively clear roles: active observation provides physical evidence about the current environment, whereas user clarification supplies intent or preference information that cannot be determined directly from visual evidence. In more general real-world interactions, however, this division of roles does not always hold. A user may possess information about an object's identity, history, or function that cannot be recovered from its current appearance, while the robot's interpretation based on visual evidence may be inconsistent with that information. Such a conflict does not necessarily imply that either the user or the visual system is incorrect; rather, different information sources may have access to different observations and prior knowledge. The robot must then address a problem beyond deciding whether to continue observing or ask the user: it must assess the reliability of different evidence sources and determine whether to observe again, request further clarification, seek another form of verification, or defer target selection when a conflict arises. This work does not address such cross-source evidence conflicts, and its conclusions therefore primarily apply to embodied disambiguation processes in which physical observations and user feedback do not exhibit substantial inconsistency. Future work could incorporate source reliability and active verification into the decision process, enabling robots to dynamically assess confidence and reconcile evidence across visual observations, user statements, and other knowledge sources.

\section{Conclusion}
\label{sec:conclusion}

This paper studied the complementary roles of physical observation information and user-intent information in embodied target disambiguation and proposed a vision-language framework with active observation as its information backbone. The robot changes its viewpoint to acquire physical evidence missing from the current observation and, on the basis of updated visual understanding, continues observing, requests user clarification, or completes target selection; the object names, labels, and semantic attributes acquired through active observation can also improve language interaction when it remains necessary. Real-robot experiments show that active perception can recover local visual evidence, expand the currently useful observation range, and provide more complete visual grounding for user clarification. These results indicate that perception and asking in embodied disambiguation are not independent information channels, but can form a continuous observation--interaction--selection process through active information acquisition.

\bibliographystyle{IEEEtran}
\bibliography{references}

@inproceedings{shridhar2018interactive,
  author    = {Mohit Shridhar and David Hsu},
  title     = {Interactive Visual Grounding of Referring Expressions for Human-Robot Interaction},
  booktitle = {Robotics: Science and Systems},
  year      = {2018},
  doi       = {10.15607/RSS.2018.XIV.028}
}

@article{shridhar2020ingress,
  author    = {Mohit Shridhar and Dixant Mittal and David Hsu},
  title     = {INGRESS: Interactive Visual Grounding of Referring Expressions},
  journal   = {The International Journal of Robotics Research},
  volume    = {39},
  number    = {2--3},
  pages     = {217--232},
  year      = {2020},
  doi       = {10.1177/0278364919897133}
}

@inproceedings{zhang2021invigorate,
  author    = {Hanbo Zhang and Yunfan Lu and Cunjun Yu and David Hsu and Xuguang Lan and Nanning Zheng},
  title     = {INVIGORATE: Interactive Visual Grounding and Grasping in Clutter},
  booktitle = {Robotics: Science and Systems},
  year      = {2021},
  doi       = {10.15607/RSS.2021.XVII.020}
}

@inproceedings{yang2022attribute,
  author    = {Yang Yang and Xibai Lou and Changhyun Choi},
  title     = {Interactive Robotic Grasping with Attribute-Guided Disambiguation},
  booktitle = {2022 IEEE International Conference on Robotics and Automation (ICRA)},
  pages     = {8914--8920},
  year      = {2022},
  doi       = {10.1109/ICRA46639.2022.9811967}
}

@inproceedings{mo2023seeask,
  author    = {Yuchen Mo and Hanbo Zhang and Tao Kong},
  title     = {Towards Open-World Interactive Disambiguation for Robotic Grasping},
  booktitle = {2023 IEEE International Conference on Robotics and Automation (ICRA)},
  pages     = {8061--8067},
  year      = {2023},
  doi       = {10.1109/ICRA48891.2023.10161333}
}

@inproceedings{ren2023knowno,
  author    = {Allen Z. Ren and Anushri Dixit and Alexandra Bodrova and Sumeet Singh and Stephen Tu and Noah Brown and Peng Xu and Leila Takayama and Fei Xia and Jake Varley and Zhenjia Xu and Dorsa Sadigh and Andy Zeng and Anirudha Majumdar},
  title     = {Robots That Ask For Help: Uncertainty Alignment for Large Language Model Planners},
  booktitle = {Proceedings of the Conference on Robot Learning},
  series    = {Proceedings of Machine Learning Research},
  volume    = {229},
  pages     = {661--682},
  year      = {2023},
  publisher = {PMLR},
  url       = {https://proceedings.mlr.press/v229/ren23a.html}
}

@article{park2024clara,
  author  = {Jeongeun Park and Seungwon Lim and Joonhyung Lee and Sangbeom Park and Minsuk Chang and Youngjae Yu and Sungjoon Choi},
  title   = {CLARA: Classifying and Disambiguating User Commands for Reliable Interactive Robotic Agents},
  journal = {IEEE Robotics and Automation Letters},
  volume  = {9},
  number  = {2},
  pages   = {1059--1066},
  year    = {2024},
  doi     = {10.1109/LRA.2023.3338514}
}

@inproceedings{kang2024prograsp,
  author    = {Gi-Cheon Kang and Junghyun Kim and Jaein Kim and Byoung-Tak Zhang},
  title     = {PROGrasp: Pragmatic Human-Robot Communication for Object Grasping},
  booktitle = {2024 IEEE International Conference on Robotics and Automation (ICRA)},
  pages     = {3304--3310},
  year      = {2024},
  doi       = {10.1109/ICRA57147.2024.10610543}
}

@inproceedings{chisari2025ambres,
  author    = {Eugenio Chisari and Jan Ole von Hartz and Fabien Despinoy and Abhinav Valada},
  title     = {Robotic Task Ambiguity Resolution via Natural Language Interaction},
  booktitle = {2025 IEEE/RSJ International Conference on Intelligent Robots and Systems (IROS)},
  pages     = {14821--14827},
  year      = {2025},
  doi       = {10.1109/IROS60139.2025.11247661}
}

@inproceedings{radford2021clip,
  author    = {Radford, Alec and Kim, Jong Wook and Hallacy, Chris and Ramesh, Aditya and Goh, Gabriel and Agarwal, Sandhini and Sastry, Girish and Askell, Amanda and Mishkin, Pamela and Clark, Jack and Krueger, Gretchen and Sutskever, Ilya},
  title     = {Learning Transferable Visual Models From Natural Language Supervision},
  booktitle = {Proceedings of the 38th International Conference on Machine Learning},
  series    = {Proceedings of Machine Learning Research},
  volume    = {139},
  pages     = {8748--8763},
  year      = {2021},
  publisher = {PMLR},
  url       = {https://proceedings.mlr.press/v139/radford21a.html}
}

@inproceedings{alayrac2022flamingo,
  author    = {Alayrac, Jean-Baptiste and Donahue, Jeff and Luc, Pauline and Miech, Antoine and Barr, Iain and Hasson, Yana and Lenc, Karel and Mensch, Arthur and Millican, Katherine and Reynolds, Malcolm and Ring, Roman and Rutherford, Eliza and Cabi, Serkan and Han, Tengda and Gong, Zhitao and Samangooei, Sina and Monteiro, Marianne and Menick, Jacob L. and Borgeaud, Sebastian and Brock, Andy and Nematzadeh, Aida and Sharifzadeh, Sahand and Binkowski, Mikolaj and Barreira, Ricardo and Vinyals, Oriol and Zisserman, Andrew and Simonyan, Karen},
  title     = {Flamingo: a Visual Language Model for Few-Shot Learning},
  booktitle = {Advances in Neural Information Processing Systems},
  volume    = {35},
  year      = {2022},
  doi       = {10.52202/068431-1723},
  url       = {https://proceedings.neurips.cc/paper_files/paper/2022/hash/960a172bc7fbf0177ccccbb411a7d800-Abstract-Conference.html}
}

@inproceedings{li2023blip2,
  author    = {Li, Junnan and Li, Dongxu and Savarese, Silvio and Hoi, Steven},
  title     = {{BLIP}-2: Bootstrapping Language-Image Pre-training with Frozen Image Encoders and Large Language Models},
  booktitle = {Proceedings of the 40th International Conference on Machine Learning},
  series    = {Proceedings of Machine Learning Research},
  volume    = {202},
  pages     = {19730--19742},
  year      = {2023},
  publisher = {PMLR},
  url       = {https://proceedings.mlr.press/v202/li23q.html}
}

@inproceedings{liu2023llava,
  author    = {Liu, Haotian and Li, Chunyuan and Wu, Qingyang and Lee, Yong Jae},
  title     = {Visual Instruction Tuning},
  booktitle = {Advances in Neural Information Processing Systems},
  volume    = {36},
  year      = {2023},
  doi       = {10.52202/075280-1516},
  url       = {https://proceedings.neurips.cc/paper_files/paper/2023/hash/6dcf277ea32ce3288914faf369fe6de0-Abstract-Conference.html}
}

@inproceedings{ichter2023saycan,
  author    = {Ichter, Brian and Brohan, Anthony and Chebotar, Yevgen and Finn, Chelsea and Hausman, Karol and Herzog, Alexander and Ho, Daniel and Ibarz, Julian and Irpan, Alex and Jang, Eric and Julian, Ryan and Kalashnikov, Dmitry and Levine, Sergey and Lu, Yao and Parada, Carolina and Rao, Kanishka and Sermanet, Pierre and Toshev, Alexander T. and Vanhoucke, Vincent and Xia, Fei and Xiao, Ted and Xu, Peng and Yan, Mengyuan and Brown, Noah and Ahn, Michael and Cortes, Omar and Sievers, Nicolas and Tan, Clayton and Xu, Sichun and Reyes, Diego and Rettinghouse, Jarek and Quiambao, Jornell and Pastor, Peter and Luu, Linda and Lee, Kuang-Huei and Kuang, Yuheng and Jesmonth, Sally and Joshi, Nikhil J. and Jeffrey, Kyle and Ruano, Rosario Jauregui and Hsu, Jasmine and Gopalakrishnan, Keerthana and David, Byron and Zeng, Andy and Fu, Chuyuan Kelly},
  title     = {Do As I Can, Not As I Say: Grounding Language in Robotic Affordances},
  booktitle = {Proceedings of The 6th Conference on Robot Learning},
  series    = {Proceedings of Machine Learning Research},
  volume    = {205},
  pages     = {287--318},
  year      = {2023},
  publisher = {PMLR},
  url       = {https://proceedings.mlr.press/v205/ichter23a.html}
}

@inproceedings{driess2023palme,
  author    = {Driess, Danny and Xia, Fei and Sajjadi, Mehdi S. M. and Lynch, Corey and Chowdhery, Aakanksha and Ichter, Brian and Wahid, Ayzaan and Tompson, Jonathan and Vuong, Quan and Yu, Tianhe and Huang, Wenlong and Chebotar, Yevgen and Sermanet, Pierre and Duckworth, Daniel and Levine, Sergey and Vanhoucke, Vincent and Hausman, Karol and Toussaint, Marc and Greff, Klaus and Zeng, Andy and Mordatch, Igor and Florence, Pete},
  title     = {{PaLM}-E: An Embodied Multimodal Language Model},
  booktitle = {Proceedings of the 40th International Conference on Machine Learning},
  series    = {Proceedings of Machine Learning Research},
  volume    = {202},
  pages     = {8469--8488},
  year      = {2023},
  publisher = {PMLR},
  url       = {https://proceedings.mlr.press/v202/driess23a.html}
}

@article{brohan2022rt1,
  author    = {Brohan, Anthony and Brown, Noah and Carbajal, Justice and Chebotar, Yevgen and Dabis, Joseph and Finn, Chelsea and Gopalakrishnan, Keerthana and Hausman, Karol and Herzog, Alex and Hsu, Jasmine and Ibarz, Julian and Ichter, Brian and Irpan, Alex and Jackson, Tomas and Jesmonth, Sally and Joshi, Nikhil J. and Julian, Ryan and Kalashnikov, Dmitry and Kuang, Yuheng and Lee, Kuang-Huei and Levine, Sergey and Lu, Yao and Malla, Utsav and Manjunath, Deeksha and Mordatch, Igor and Nachum, Ofir and Parada, Carolina and Peralta, Jodilyn and Perez, Emily and Pertsch, Karl and Quiambao, Jornell and Rao, Kanishka and Ryoo, Michael and Sayed, Kevin and Singh, Jaspiar and Sontakke, Sumedh and Stone, Austin and Tan, Clayton and Tran, Huong and Vanhoucke, Vincent and Vega, Steve and Vuong, Quan and Xia, Fei and Xiao, Ted and Xu, Peng and Xu, Tianhe and Zitkovich, Brianna},
  title     = {RT-1: Robotics Transformer for Real-World Control at Scale},
  journal   = {arXiv preprint arXiv:2212.06817},
  year      = {2022},
  url       = {https://arxiv.org/abs/2212.06817}
}

@inproceedings{zitkovich2023rt2,
  author    = {Zitkovich, Brianna and Yu, Tianhe and Xu, Sichun and Xu, Peng and Xiao, Ted and Xia, Fei and Wu, Jialin and Wohlhart, Paul and Welker, Stefan and Wahid, Ayzaan and Vuong, Quan and Vanhoucke, Vincent and Tran, Huong and Soricut, Radu and Singh, Anikait and Singh, Jaspiar and Sermanet, Pierre and Sanketi, Pannag R. and Salazar, Grecia and Ryoo, Michael S. and Reymann, Krista and Rao, Kanishka and Pertsch, Karl and Mordatch, Igor and Michalewski, Henryk and Lu, Yao and Levine, Sergey and Lee, Lisa and Lee, Tsang-Wei Edward and Leal, Isabel and Kuang, Yuheng and Kalashnikov, Dmitry and Julian, Ryan and Joshi, Nikhil J. and Irpan, Alex and Ichter, Brian and Hsu, Jasmine and Herzog, Alexander and Hausman, Karol and Gopalakrishnan, Keerthana and Fu, Chuyuan and Florence, Pete and Finn, Chelsea and Dubey, Kumar Avinava and Driess, Danny and Ding, Tianli and Choromanski, Krzysztof Marcin and Chen, Xi and Chebotar, Yevgen and Carbajal, Justice and Brown, Noah and Brohan, Anthony and Arenas, Montserrat Gonzalez and Han, Kehang},
  title     = {RT-2: Vision-Language-Action Models Transfer Web Knowledge to Robotic Control},
  booktitle = {Proceedings of The 7th Conference on Robot Learning},
  series    = {Proceedings of Machine Learning Research},
  volume    = {229},
  pages     = {2165--2183},
  year      = {2023},
  publisher = {PMLR},
  url       = {https://proceedings.mlr.press/v229/zitkovich23a.html}
}

@inproceedings{openx2024rtx,
  author    = {{Open X-Embodiment Collaboration}},
  title     = {Open X-Embodiment: Robotic Learning Datasets and RT-X Models},
  booktitle = {2024 IEEE International Conference on Robotics and Automation (ICRA)},
  pages     = {6892--6903},
  year      = {2024},
  doi       = {10.1109/ICRA57147.2024.10611477},
  url       = {https://robotics-transformer-x.github.io/}
}

@inproceedings{kim2024openvla,
  author    = {Kim, Moo Jin and Pertsch, Karl and Karamcheti, Siddharth and Xiao, Ted and Balakrishna, Ashwin and Nair, Suraj and Rafailov, Rafael and Foster, Ethan P. and Sanketi, Pannag R. and Vuong, Quan and Kollar, Thomas and Burchfiel, Benjamin and Tedrake, Russ and Sadigh, Dorsa and Levine, Sergey and Liang, Percy and Finn, Chelsea},
  title     = {OpenVLA: An Open-Source Vision-Language-Action Model},
  booktitle = {Proceedings of The 8th Conference on Robot Learning},
  series    = {Proceedings of Machine Learning Research},
  volume    = {270},
  pages     = {2679--2713},
  year      = {2024},
  url       = {https://openvla.github.io/}
}

@inproceedings{octo2024generalist,
  author    = {{Octo Model Team} and Ghosh, Dibya and Walke, Homer and Pertsch, Karl and Black, Kevin and Mees, Oier and Dasari, Sudeep and Hejna, Joey and Xu, Charles and Luo, Jianlan and Kreiman, Tobias and {You Liang} Tan and Chen, Lawrence Yunliang and Sanketi, Pannag and Vuong, Quan and Xiao, Ted and Sadigh, Dorsa and Finn, Chelsea and Levine, Sergey},
  title     = {Octo: An Open-Source Generalist Robot Policy},
  booktitle = {Proceedings of Robotics: Science and Systems},
  address   = {Delft, The Netherlands},
  year      = {2024},
  url       = {https://octo-models.github.io/}
}

\end{CJK*}
\end{document}